\documentclass{article}

\usepackage{arxiv}

\usepackage[utf8]{inputenc} 
\usepackage[T1]{fontenc}    
\usepackage{hyperref}       
\usepackage{url}            
\usepackage{booktabs}       
\usepackage{amsfonts}       
\usepackage{nicefrac}       
\usepackage{microtype}      
\usepackage{lipsum}		
\usepackage{graphicx}
\usepackage[square,numbers,sort]{natbib}
\usepackage{doi}

\usepackage[cmex10]{amsmath}
\usepackage{amssymb}
\usepackage{algorithmic}

\title{Index $t$-SNE: Tracking Dynamics of High-Dimensional Datasets with Coherent Embeddings.}

\author{
	\href{https://orcid.org/0000-0002-9072-1535}{\includegraphics[scale=0.06]{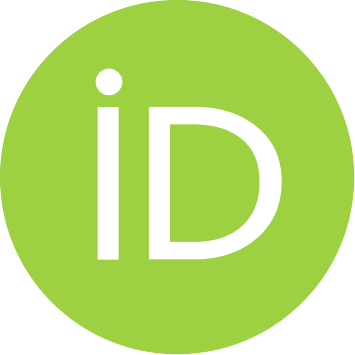}\hspace{1mm}Ga\"elle Candel} \\
	Wordline TSS Labs, Paris \\
	\texttt{firstname.lastname@worldline.com} \\
	\& \\
	D\'epartement d'informatique de l'ENS \\
	ENS, CNRS, PSL University, Paris \\
	\texttt{firstname.lastname@ens.fr}\\
	\And
  David Naccache \\
	D\'epartement d'informatique de l'ENS \\
	ENS, CNRS, PSL University, Paris \\
	\texttt{firstname.lastname@ens.fr}\\
}

\date{}

\renewcommand{\headeright}{}
\renewcommand{\undertitle}{}
\renewcommand{\shorttitle}{Index $t$-SNE}

\hypersetup{
pdftitle={Index $t$-SNE: Tracking Dynamics of High-Dimensional Datasets with Coherent Embeddings.},
pdfsubject={cs.AI, cS.LG},
pdfauthor={Ga\"elle Candel, Davif Naccache},
pdfkeywords={Citation graphs, Concept drift, Data visualization, Dimension reduction, Embedding, Monitoring, Reusability, $t$-SNE, Unsupervised learning},
}

\begin{document}
\maketitle

\begin{abstract}
	$t$-SNE is an embedding method that the data science community has widely used.
	Two interesting characteristics of t-SNE are the structure preservation property and the answer to the crowding problem, where all neighbors in high dimensional space cannot be represented correctly in low dimensional space.
	$t$-SNE preserves the local neighborhood, and similar items are nicely spaced by adjusting to the local density.
	These two characteristics produce a meaningful representation, where the cluster area is proportional to its size in number,
	and relationships between clusters are materialized by closeness on the embedding.

	This algorithm is non-parametric, therefore two initializations of the algorithm would lead to two different embedding.
	In a forensic approach, analysts would like to compare two or more datasets using their embedding.
	An approach would be to learn a parametric model over an embedding built with a subset of data.
	While this approach is highly scalable, points could be mapped at the same exact position, making them indistinguishable.
	This type of model would be unable to adapt to new outliers nor concept drift.

	This paper presents a methodology to reuse an embedding to create a new one, where cluster positions are preserved.
	The optimization process minimizes two costs, one relative to the embedding shape and the second relative to the support embedding' match.
	The proposed algorithm has the same complexity than the original $t$-SNE to embed new items, and a lower one when considering the embedding of a dataset sliced into sub-pieces.
	The method showed promising results on a real-world dataset, allowing to observe the birth, evolution and death of clusters.
	The proposed approach facilitates identifying significant trends and changes, which empowers the monitoring high dimensional datasets' dynamics.
\end{abstract}


\section{Introduction}

High dimensional datasets are very rich sources of information. Because of their wideness, they are difficult to investigate, process, and represent in a simple manner. An appropriate reduction of width would benefit from:
\begin{itemize}
  \item Storage cost reduction;
  \item Computational cost reduction;
  \item Denoising / Information compression;
  \item Synthetic data visualization.
\end{itemize}

For a dataset $X \in \mathbb{R}^{n \times d_0}$, with $n$ elements and $d_0$ dimensions, a reduction into $Y \in \mathbb{R}^{n \times d_1}$ offers a reduction of the memory footprint of $100 (1 - \frac{d_1}{d_0})$ \%, which is non-negligible for large datasets.
For linear algorithms, the computational cost is reduced by the same factor.
The core information in those wide datasets is hard to identify because of the large number of features and correlation, and redundancy. The use of methods that condense the information enables to obtain a synthetic view of the dataset,
 which would benefit post-processing algorithms or data analysts' work.
The synthetic view can also be used to display results by coloring items according to their predicted value.

A dimension reduction can be performed following different approaches:
\begin{itemize}
  \item Automatic feature selection;
  \item Human engineered feature;
  \item Automatic feature extraction.
\end{itemize}
Automatic feature selection selects some features among all available according to some characteristics. The feature importance can be evaluated using Shapley values \cite{feature_sel_shapley_value} or removing redundant features \cite{hall1999correlationbased}.
These approaches are relatively straightforward to put into practice.
Nonetheless, some information may be lost if the signal is too weak, or if the number of selected features is too small.

Feature engineering allows shaping human knowledge into an algorithmic form.
This approach is practical when a minimal amount of data is available, preventing the use of automatic algorithms. It is necessary in specific cases to transform human data into processable information. Dates are a good example: a computer cannot understand directly that $\mathbf{x}=[28, 2]$ and $\mathbf{y}=[1, 3]$ represent dates and that $\|\mathbf{x} - \mathbf{y}\| = 1 \text{ or } 2\}$ days, depending on this is a leap year or not.
Apart from this example with an exact answer, there is no guarantee if the transformation would help or prevent later algorithms from performing good predictions. As these features are not learned from data, they are likely to be stable, i.e., not tricked by outliers which could perturb the learning. They offer some form of \textit{explainability}, as the engineer can describe the meaning of the transformation in the human language.
In the absence of previous knowledge, their development is costly in operating time, and on a large dataset, an expert may miss some essential features.

Automatic feature extraction replaces handcrafted features with algorithm-learned features. This is a broad research area, with statistical compression methods such as Principal Component Analysis \cite{PCA}, or methods based on deep neural networks, like autoencoders \cite{SCAE_HFE}, \cite{MAGGIPINTO2018126}.

Automatic feature extraction can be decomposed into two categories, based on reusability on new data:
\begin{itemize}
  \item Parametric methods;
  \item Non-parametric methods.
\end{itemize}

Parametric methods, such as PCA \cite{PCA}, Self-Organizing Maps \cite{kohonen-self-organizing-maps-2001}, learn a mapping function $f: X \rightarrow Y$, which minimize a quantity of the form $\arg_f \min Cost(X, f(X))$.
When learned, the function $f$ can be reused on any new input $X'$.
The inference time of most of these methods is linear, such as $f(X) = [f(\mathbf{x}_1), f(\mathbf{x}_2), ..., f(\mathbf{x}_n)]$.
For a large input $X$, the computation can be distributed on several machines, which enhance scalability to large datasets. In contrast, non-parametric methods minimize a specific quantity $\min_Y Cost(X, Y)$ directly by optimizing $Y$'s values.
No function is learned, which prevents the reusability of a previous computation for new data. This is the case of ISOMAP \cite{tenenbaum_global_2000_isomap}, UMAP \cite{UMAPmcinnes2018uniform} or $t$-SNE \cite{vanDerMaaten2008}.
Despite the non-reusability, the two most recent methods, $t$-SNE and UMAP, have been extensively used by the machine learning community.
Those methods have been successful at representing high dimensional datasets, by adapting to disparate scaling and non-homogeneous densities while letting appear  clustering structures.

The strength of $t$-SNE comes from its ability to deal with  heterogeneous scaling and the crowding problem. In high dimensional space, an item may have many neighbors around, all distant from each other. By reducing the number of dimensions, it is impossible to preserve the distance between an item and its neighbors and between the neighbors.
If distance to the item is preserved, neighbours' distance will decrease, making them closer than in high dimensional space.
This corresponds to the crowding problem.

Instead of preserving all the distances, $t$-SNE preserves it locally. The algorithm adapts for each item to the local density, taking into account a small group of neighbors.
This local adaptation makes the power of $t$-SNE, as the points in a very dense cluster are separated from each other. Consequently, the number of items in a cluster is proportional to its visual area on the embedding, which helps an analyst look at the dataset composition.
The other impact of the local adaptation is on the ability to deal with heterogeneous data scaling. All items fit together on the same visual space regardless their initial distance to the dataset's mean. These characteristics lead to embeddings with excellent visual qualities.

Despite the high quality of the obtained embeddings, $t$-SNE is a non-parametric method, where no function is learned.
The outcome of running $t$-SNE twice with the same input $X$ leads to two different embeddings $Y^{(0)}$ and $Y^{(1)}$, with no equivalence between the positions. This is due to the initialization, which starts with randomly generated vectors.

The initialisation process can be controlled to improve determinism. Many works such as \cite{Kobak453449} proposed to initialize the embedding with PCA coefficients.
Starting with these positions improves the repeatability, but does not ensure the regeneration of large scale structures for different datasets, as $t$-SNE preserves local neighborhood only.
The work of \cite{Kobak453449} proposes a method to create large scale structures using two $t$-SNE steps, which enables to obtain embeddings with large and low scale similarities. The algorithm starts with the PCA coefficients as initial item positions, followed by a $t$-SNE step adjusted to take into account far-range neighborhood.
These positions are reused by another $t$-SNE step, taking into account small-range neighborhood, letting appear a finer structure. This work was successful for visual analytics, allowing the preservation of relative cluster positions over multiple datasets.
The embeddings are visually similar, but the preservation of cluster positions is not exact, preventing the use of the same algorithm on all embeddings.

A naive approach to compare two datasets using their embedding is to compute the joint embedding over the consolidated dataset $X'' = [X, X']$.
There are two limitations to this approach. The first one concerns the computational cost, as the complexity of $t$-SNE is in $\mathcal{O}(n^2)$ for the worst case.
The work of \cite{DBLP:journals/corr/abs-1805-10817} proposes an approximation of the different forces, claiming a linear complexity. Nonetheless, the second limitation concerns the data availability. If the two datasets are available now, but a third would arrive later, the embedding corresponding to $[X^{(0)}, X^{(1)}]$ would not share spatial correspondences with the embedding obtained with $[X^{(1)}, X^{(2)}]$.

To obtain consistency in the item positioning, several works \cite{pmlr-v5-maaten09a},\cite{DBLP:journals/corr/abs-1710-05128} proposed the use of deep neural networks to mimic the behavior of $t$-SNE.
As with any trained algorithm with no memory nor update mechanism, the inference results is purely deterministic. The algorithm would be able to map correctly a dataset with a distribution similar to the training dataset. However, for a dataset with a different distribution, the model would not adapt to the new density, leading to overcrowded and/or depleted areas.
The neural network would not be able to adapt to concept drift, nor as new outliers as these models' generalizability is limited to their training set.

Instead of learning how to make an embedding, LION $t$-SNE \cite{DBLP:journals/corr/abs-1708-04983} proposed an answer to \textit{where new points should be put}, taking into account the points already present and the empty areas left.
New points are positioned nearby their nearest neighbors without moving existing items from their location. Items that do not have relevant neighbors in the input space are positioned on an empty area of the embedding, filled later with more relevant neighbors.
This approach allows adding a few points on the previous embedding, keeping the possible visual quality of the embedding. While this work deals correctly with item's addition, normal or outlier, it does not deal with update nor deletion.
Last point concerns the scalability. The addition of a few points is likely to preserve the general aspect of the embedding.
However, the shape of the resulting embedding after a massive addition of items is unknown.

Last work to mention is Dynamic t-SNE \cite{10.5555/3058878.3058894} which updates the embedding $Y^{(t)}$ into $Y^{(t+1)}$ after a change from $X^{(t)}$ into $X^{(t+1)}$.
It assumes that there is a one-to-one correspondence between items of $X^{(t)}$ and $X^{(t+1)}$. This setup corresponds to a monitoring situation, where the data coming from a fixed number of sensors arrives at each time step. To compute $Y^{(t+1)}$, $dt$-SNE starts with the previous embedding positions $Y^{(t)}$, and tries to minimize the cost defined by $t$-SNE relatively to $X^{(t+1)}$.
A penalty is added on the displacement of $Y^{(t+1)}$ from the initial position, which enables to keep the embedding coherent over multiple time steps. While this work addresses the updatability, as no items can be added nor deleted, the usability is restricted to particular use-cases, such as multidimensional time series.

In this paper, we abort the problem of the reusability of a $t$-SNE embedding.
The proposed approach is inspired by $dt$-SNE, concerning the idea of using the previous embedding as a support to the new embedding.
The support embedding is not used to initialize the new item positions but to guide them towards neighbors location.
Compared to $dt$-SNE, the scope is broadened because there is no constraint on the integrated elements, nor on their number or distribution.
By enlarging to the addition, updating and deleting items, our method can be used in many more real-world monitoring situations,
such as when some sensors are added to the system or removed due to failure.
The approach is not limited to the temporal dataset, but to any \textit{index} variable, a discrete or continuous variable, such as temperature or speed.
Dataset can be sorted according to this variable, and successive embeddings can be issued to track the impact of the \textit{index} variable over the data distribution. In other words, it allows to obtain embedding conditional to the index variable of interest.nThe method is called \textit{index} $t$-SNE, abbreviated $it$-SNE, for this reason.

In the first section, the main equations governing $t$-SNE optimization process are introduced. This section is followed by the description of $it$-SNE, reusing part of the initial $t$-SNE scheme. Then, the methods section describe the different datasets and evaluation metrics, followed by the experimental results.
Last, this article finishes with a discussion followed by its conclusion.

\section{$t$-SNE Formulation}
$t$-SNE \cite{vanDerMaaten2008} is a structure-preserving embedding algorithm
trying to preserve the local neighborhood of items in a low dimensional space.
Given a dataset $X \in \mathbb{R}^{(n \times d)}$, of $n$ items lying in a $d$ dimensional space,
the goal is to generate its corresponding embedding $Y \in \mathbb{R}^{(n \times d_e)}$:

$$Y \leftarrow t\text{-SNE}(X; d_e, perp)$$
where  $d_e$ is the number of embedding dimensions, often set to $2$, and $perp$ is the perplexity parameter.
Two items $i$ and $j$ neighbors in $X$ must be neighbors in $Y$. The definition of \textit{neighbors} depends of two things: the local density around an item,
and the user-defined perplexity parameter which represents the average number of neighbors to consider.
Rather than reasoning in terms of distances, the algorithm uses probabilities, computed from pairwise distances, to optimize $Y$.

\subsection{Interaction Probability}

$t$-SNE tries to adapt the embedding vector $Y$ to $X$ using their respective probability matrices $P$ and $Q$, both of dimension $n \times n$.
These probabilities represent the degree of relatedness of two items in their respective space.
A large probability  corresponds to a high proximity, while a smaller to a large distance.
The input and output probability matrices are computed differently to create a small asymmetry.

\subsubsection{Input Probability Matrix}
The conditional probability of item $j$ with respect to $i$ is defined as:

\begin{equation} \label{eq:input_proba}
  p_{j | i} = \frac{1}{Z_i}\exp \left(- \frac{\|\mathbf{x}_i - \mathbf{x}_j\|^2}{2 \sigma_i^2} \right)
\end{equation}

By convention, $p_{i | i} = 0$ as an item does not interact with itself, and $Z_i = \sum_{j\neq i} \exp \left(- \frac{\|\mathbf{x}_i - \mathbf{x}_j\|^2}{2 \sigma_i^2} \right)$ is the normalization constant of item $i$, such as $\sum_j p_{j | i} = 1$.

The standard deviation parameter $\sigma_i$ adapts the kernel range to the local density around item $i$.
The optimal value of $\sigma_i$ is obtained by binary search to match the \textit{perplexity}.
The perplexity is a user-defined parameter that represents the average number of neighbors of an item, defined formally as:

$$Perp(P_i) = 2^{H(P_i)}$$
where $H(P_i)$ is the Shannon entropy calculated as:

$$H(P_i) = - \sum_{j \neq i} p_{j | i} \log_2(p_{j | i})$$

The joint probability between $i$ and $j$ is defined as $p_{i,j} = \frac{p_{i | j} + p_{j | i}}{2 n}$.
These equations enable the computation of the symmetric probability matrix $P$ given a particular dataset $X$ and a perplexity target.

\subsubsection{Output Probability Matrix}
The output probability matrix $Q$ is obtained in a similar manner using the embedding vector $Y$.
As the goal is to obtain homogeneous distances between neighbors, there is no adaptation to local neighborhood. Another difference concerns the kernel choice. Instead of an exponential kernel, a $t$-student kernel with one degree of freedom is used. This kernel asymmetry allows modifying the long-range interactions, which leads to repulsive forces between non-neighbors items.

The joint probability between item $i$ and $j$ is calculated as:

\begin{equation} \label{eq:output_proba}
  q_{i, j} = \frac{1}{V}(1 + \|\mathbf{y}_i - \mathbf{y}_j\|^2)^{-1}
\end{equation}
with $q_{i, i} = 0$ and $V = \sum_{k \neq \ell} (1 + \|\mathbf{y}_k - \mathbf{y}_{\ell}\|^2)^{-1}$ the global normalization constant, which lead to $\sum_{i, j} q_{i, j} = 1$.

\subsection{Cost Minimization}
The dissimilarity between the two probability matrices $P$ and $Q$ is measured using the Kullback-Leibler divergence:

\begin{equation} \label{eq:KL}
  KL(P \| Q) = \sum_{i} C_i = \sum_i \sum_{j \neq i} p_{i, j} \log \frac{p_{i, j}}{q_{i, j}}
\end{equation}
where $C_i$ the cost associated to item $i$.

By deriving \eqref{eq:KL}, a simple form of the gradient is obtained:

\begin{equation} \label{eq:derivative_tsne}
  \frac{\partial C_i}{\partial y_i} = 4 \sum_j (p_{i, j} - q_{i, j}) \frac{\mathbf{y}_i - \mathbf{y}_j}{1 + \|\mathbf{y}_i - \mathbf{y}_j\|^2}
\end{equation}

The algorithm uses the gradient descent approach to minimize the cost by updating the initial $Y(0)$ solution:

\begin{equation} \label{eq:gradient_descent_momentum}
  \begin{array}{ll}
    Y(t) = & Y(t-1) - \alpha(t) \frac{\partial C}{\partial Y}(t) \\
          & + \eta(t) \left(Y(t-1) - Y(t-2) \right) \\
    \end{array}
\end{equation}
with $\alpha(t)$ the learning rate, adjusted over time, and $\eta(t)$ the momentum rate.
The matrix $Q$ is recomputed at each update step according to the newly obtained $Y(t)$, while the matrix $P$ is not as the input vector is left unchanged. The computation of $Q$ at each step is the most costly operation, which leads to a complexity in $\mathcal{O}(n^2)$ without optimization.

In the original paper \cite{vanDerMaaten2008}, many optimization tricks are used.
For instance, \textit{early exaggeration} replaces temporary $(p_{i, j} - q_{i, j})$ by $(k p_{i, j} - q_{i, j})$, where $k > 1$. This trick amplifies the  attraction forces between nearest neighbors, which fasten the formation of separated clusters.
After some steps, the factor is set back to $k=1$ which lets nearest neighbors to separate from each others.

Another trick is to add Gaussian noise of small amplitude to the gradient to get out of local minima at start.

The last point to be made is the $\alpha(t)$ learning rate. This rate is updated at each step to boost it in the right directions and slow it down in uncertain situations.

\section{Indexed $t$-SNE}

The initial formulation of $t$-SNE leads to a different embedding for each new initialization.
Instead of learning a new embedding from scratch, $it$-SNE takes advantage of prior embedding to optimize the new embedding.

Given a support dataset $X^{(0)} \in \mathbb{R}^{n_0 \times d}$ of $n_0$ items and its corresponding embedding $Y^{(0)} \in \mathbb{R}^{n_0 \times d_e}$,
the goal is to generate an embedding $Y^{(1)} \in \mathbb{R}^{n_1 \times d_e}$ corresponding to $X^{(1)} \in \mathbb{R}^{n_1 \times d}$ of $n_1$ items.

$$Y^{(1)} \leftarrow it\text{-SNE}(X^{(1)}, X^{(0)}, Y^{(0)}; perp)$$

Two items $i$ and $j$ neighbors in the input space must be neighbors in the embedding space,
regardeless of their origin dataset.

\subsection{Cost}

To achieve this goal, $it$-SNE minimizes two independent costs:

\begin{itemize}
  \item the \textit{intra} cost, $C^{(1)}$, defined as in $t$-SNE using $(X^{(1)}, Y^{(1)})$;
  \item the \textit{inter} cost, $C^{(0, 1)}$, corresponding to joint interactions between $(X^{(0)}, Y^{(0)})$ and $(X^{(1)}, Y^{(1)})$.
\end{itemize}

The total cost to minimize is:

\begin{equation} \label{eq:total_cost}
  C_{tot}^{(1)} = C^{(1)} + C^{(0, 1)}
\end{equation}

\subsection{Interaction Probability}

Similar to $t$-SNE, the probability matrices are defined to represent items relationships.
$P^{(0, 1)}$ denotes the probability matrix between input data $X^{(0)}$ and $X^{(1)}$,  and
$Q^{(0, 1)}$ for their respective embeddings $Y^{(0)}$ and $Y^{(1)}$.

\subsubsection{Input Interaction}
For two items $\mathbf{x}_i \in X^{(0)}$ and $\mathbf{x}_j \in X^{(1)}$, the input probability is defined as:

\begin{equation} \label{eq:inter_input_proba}
  p^{(0, 1)}_{i | j} = \frac{1}{Z'_j} \exp \left(- \frac{\|\mathbf{x}_i - \mathbf{x}_j\|^2}{2 \sigma_j^2}\right)
\end{equation}
where $\sigma_j$ corresponds to the optimal parameter for $j$ obtained with $t$-SNE on $X^{(1)}$, and $Z'_j = \sum_{i} \exp \left(- \frac{\|\mathbf{x}_i - \mathbf{x}_j\|^2}{2 \sigma_j^2}\right)$ is the normalization constant to obtain $\sum_i p^{(0, 1)}_{i | j} = 1$.

The joint probability between $i$ and $j$ is defined as:

\begin{equation} \label{eq:inter_input_proba_joint}
  p^{(0, 1)}_{i, j} = p^{(1, 0)}_{j, i} = \frac{1}{2}\left(\frac{p^{(0, 1)}_{i | j}}{n_1} + \frac{p^{(1, 0)}_{j | i}}{n_0} \right)
\end{equation}

The symmetrization allows taking into account the density of the two datasets on a given location.
Additionally, the normalization allows to equalize the dataset influence.
If one dataset is larger than the other, it would contribute more are the total forces from each of its  items would be larger than for the smaller dataset.
The normalization by dataset size allow to obtain equivalent contribution.

\subsubsection{Output Interaction Probabilities}
The goal of $it$-SNE is not to place new items $Y^{(1)}$ on existing holes of $Y^{(0)}$, but to have $Y^{(1)}$ on a parallel layer of $Y^{(0)}$.
To relax forces, and to take into account embedding separation, a penalty factor $\epsilon$ is introduced to artificialy separate points belonging to different embeddings. It could be seen as new embedding dimension $d_e+1$, such as $\mathbf{y}_i^{(0)} = [y^{(0)}_{i, 1}, y^{(0)}_{i, 2}, ..., y^{(0)}_{i, d_e}, 0]$ and $\mathbf{y}_j^{(1)} = [y^{(1)}_{j, 1}, y^{(1)}_{j, 2}, ..., y^{(1)}_{j, d_e}, \epsilon]$.
The distance between two items is then $\|\mathbf{y}_i^{(0)} - \mathbf{y}_j^{(1)}\| = \|\mathbf{y}_i - \mathbf{y}_j\| + \epsilon^2$.

The kernel used for the definition of output probabilities is kept unchanged, up to the addition of $\epsilon$:

\begin{equation} \label{eq:inter_output_proba}
  q^{(0, 1)}_{i, j} = \frac{1}{V'} \left(1  + \|\mathbf{y}_i - \mathbf{y}_j\|^2 + \epsilon^2\right)^{-1}
\end{equation}
where $V' = \sum_{i, j} \left(1  + \|\mathbf{y}_i - \mathbf{y}_j\|^2 + \epsilon^2\right)^{-1}$ is the normalization constant,
which ensures $\sum_{i, j} q^{(0, 1)}_{i, j} = 1$.

\subsubsection{Cost Minimization}

The modification of $t$-SNE algorithm has a limited impact on the cost derivative formulation.
The only change impacts the strength of the gradient by the addition of the term $\epsilon^2$:

\begin{equation} \label{eq:derivative_itsne}
  \frac{\partial C^{(0, 1)}}{\partial \mathbf{y}^{(1)}_i} = 4 \sum_j (p^{(0, 1)}_{i, j} - q^{(0, 1)}_{i, j}) \frac{\mathbf{y}_j - \mathbf{y}_i}{1  + \|\mathbf{y}_i -  \mathbf{y}_j\|^2 + \epsilon^2}
\end{equation}

The initial solution $Y^{(1)}(0)$ is optimized following equation \eqref{eq:gradient_descent_momentum},
replacing $\frac{\partial C}{\partial Y}$ by $\frac{\partial C^{(1)}}{\partial Y^{(1)}} + \frac{\partial C^{(0, 1)}}{\partial Y^{(1)}}$.

\section{Experimental Setup}
\subsection{Algorithm Parametrization}

For all experiments, the target perplexity is set to $30$. The initial vector of $Y$, used to start the optimization process is drawn from a Gaussian distribution $\mathcal{N}(0, \sigma^2)$ with $\sigma=10^{-4}$.
At each iteration step, a gaussian noise of standard deviation $\sigma=10^{-4}$ is added to the gradient.

The initial learning rate $\alpha(0) = 10$ is adapted at each timestep for each item $i$ and dimension $k$ according to the similarity between the gradient and the previous displacement direction $\delta_{i, k}(t) = - \frac{\partial C}{\partial y_{i, k}}(t) . (y_{i, k}(t-1) - y_{i, k}(t-2))$:

$$\alpha_{i, k}(t) = \left\{
   \begin{array}{lll}
       \alpha_{i, k}(t-1) + 0.2 & \mbox{if }  & \delta_{i, k}(t) > 0 \\
       \alpha_{i, k}(t-1) \times 0.8 & \mbox{else.}
   \end{array}
\right.
$$

The momentum rate $\eta(t)$ is adapted over learning, such as $\eta(t) = 0.5 \text{ if } t < 250 \text{ else } 0.8$.

For $t$-SNE, the early exaggeration trick is used for the $100$ first steps, with an exaggeration factor of $2$.
For $it$-SNE, the early exaggeration was disabled.

The number of training steps for a $t$-SNE embedding and $it$-SNE is fixed to $300$ and $200$ respectively.
Unless specified, $\epsilon = 1$.

$t$-SNE and $it$-SNE were implemented in Python, using NumPy library \cite{harris2020array}.

\subsection{Datasets}
We propose to illustrate the results $it$-SNE over two types of datasets:
\begin{itemize}
  \item A synthetic dataset, where all parameters can be controlled at ease;
  \item A real-world dataset to look at its capabilities on unknown distributions.
\end{itemize}

\subsubsection{High Dimensional Gaussians}

High dimensional Gaussians are good study candidates, as all dimension are equivalent, preventing the use of renormalization methods.
The dataset was constructed inspired from the protocol described in \cite{10.5555/3058878.3058894}.

A dataset of $100$ dimensions is created by generating Gaussian clusters. $10$ Gaussian center positions $\{\pmb{\mu}_g\}_{g=1:10}$ are generated uniformly at random in $\pmb{\mu}_g \in [-0.5, 0.5]^{100}$.
For each Gaussian $g$, $100$ items are sampled from the multivariate normal distribution of mean $\pmb{\mu}_g$ and standard deviation $\sigma$.
This process leads to a dataset of $1000$ items.

In the paper Dynamic $t$-SNE \cite{10.5555/3058878.3058894}, the authors proposed to build a temporal dataset made of shrinking Gaussians.
For each item $\mathbf{x}$ associated with the Gaussian $g$, the distance between the item and the center is reduced by $10$ \% at each timestep, such as $\|\mathbf{x}(t) - \pmb{\mu}_g\| = 0.9^t\|\mathbf{x}(0) - \pmb{\mu}_g\|$.
Mechanically, $\sigma(t)$ is reduced in the same proportion, such as $\sigma(t) = 0.9^t \sigma(0)$. For this experiment, $\sigma(0) = 1.0$. The shrinking process is followed for $9$ steps, which leads to $\sigma(10) \approx 0.4$.

By changing the time direction, a similar experiment with growing Gaussians is generated, starting with $\sigma(0)=0.4$, and growing at the rate of $\sigma(t) = 0.9^{-t} \sigma(0)$.

While these two introductive experiments keep the total number of point stable and homogeneous density for each Gaussian,
we propose to build a temporal dataset where heterogeneity appears over time.
For each Gaussian $g$, an expansion rate is sampled uniformly at random from $r_g \in [0.5, 1.5]$, where $1.$ leads to an invariance of the element number.
In contrast, $0.5$ leads to a $50$ \% step reduction in the number of elements.
Each Gaussian starts with $n_g(0) = 100$ points. The number of points for Gaussian $g$ at step $t$ is $n_g(t) = \lfloor n(0) r_g^t \rceil$, but the standard deviation is kept stable with $\sigma = 0.4$.
A temporal dataset is generated for $4$ successive update steps.

\subsubsection{Citation Graphs}

A citation graph  $G = (V, E)$ is a directed acyclic graph (DAG). $V$ represents the a set of documents, like scientific papers, patents, law articles, blog posts, which are supposedly immutable.
$E$ is the set of directed edges, with $e = (a, b) \in E$ meaning that the document $a$ is referring to document $b$, implicitly but assuming that $a$ is newer than $b$.

Graphs are data structure that are difficult to represent in 2D, because they of their sparsity
and the distribution in power-law of their node degree, which leads to a small number of strongly connected nodes, and a large number of weakly connected nodes.
Nonetheless, citation graphs, as well as other real-world graphs, organise into local communities which are interesting to study.

We propose to study the evolution of research communities over time by embedding the documents published each year.
The embedding obtained for year $t$ is reused for year $t+1$, which would be used in turn to build the following embedding.
The DBLP dataset version 12 \cite{AMINER_10.1145/1401890.1402008} was used for this purpose.
This dataset corresponds to the citation graph of scientific papers around the topic of computer science.
It contains 4.894.081 papers and 45.564.149 citing relationships.
Metadata are available for the majority of the documents, providing information such as title, publication date, abstract, authors information, conference or journal reference, reference links, and keywords.
Keywords also called \textit{field of study}, are automatically extracted according to the method described in \cite{MOG_10.3389/fdata.2019.00045}.

\paragraph{Graph preprocessing}

A preprocessing removes all existing cycles, as some papers are updated after the official publication date, adding a few citations.
This phenomenon concerns a minority of papers, but creates undesirable loops.
The cycles are removed using a DFS approach, removing any edges accessed twice by a DFS branch. The main connected component is then kept, removing all papers with no bibliography or belonging to an isolated community.

\paragraph{Node Sampling}

The citation graph considered is too large to be processed at once. We took advantage of the keywords to select a subset of papers related to a particular topic.
We selected all documents related to \textit{cryptography} and related topics, which corresponded to around 100.000 documents published between 1953 and 2020.
The documents with less than one reference and less than one citation were removed, which left 70.000 documents published between 1953 and 2020,
with the majority published between 2005 and later.

\paragraph{Extracting Distance Matrix from a Citation Graph}

A graph cannot be converted to tabular data used by $t$-SNE. Nonetheless, $t$-SNE and $it$-SNE use the distance between items and do not focus on particular features.
Even if it is not possible to measure the euclidean distance between nodes on a graph, a distance matrix can be obtained.

The distance between nodes is not a great distance measure. The possible integer values are coarse measures, and the distance is not correctly defined for all pairs in a DAG.
Plus, a document may quote unrelated documents from another discipline for illustrating its argumentation with other scientific views. The node could be at distance $2$ of many papers on a completely unrelated field, just because of a single example.

A way to measure the document similarity is through bibliographic coupling \cite{BC_WEINBERG1974189}.
Two documents sharing some of their references are coupled, even if there is no direct path from one to the other in the DAG.
The strength of the coupling depends on the overlap size. As the graph is sparse, we extend the bibliographic coupling to indirect reference until distance $3$.
This strategy reduces the sparsity and improves the sensibility of the coupling.

The traditional bibliographic coupling exploits reference sets overlap to measure similarity between two papers.
Because of the graph sparsity, this method is unstable if papers make very few references.
This issue can be avoided by looking at direct and indirect references, weighting them based on their reachability.
We use the method presented in \cite{candel2021generating} to measure papers similarity, using the references up to distance $3$ and comparing two papers using the weighted cosine similarity.
The exploration of indirect neighbors reduces the sparsity and refines the relationships' strength.

This method leads to similarity values $s \in [0, 1]$, which are converted into distance by inversion:

\begin{equation} \label{eq:dist_inv}
  d = \frac{1}{s + \xi} - \frac{1}{1+\xi}
\end{equation}
where $\xi=10^{-5}$ is a small constant, which avoids division by zero and limit the maximal distance to $10^5$.
Using the deep weighted bibliographic method of \cite{candel2021generating} and \eqref{eq:dist_inv}, the citation graph can be appropriately transformed for the embedding algorithms.

\subsection{Metrics}

\subsubsection{Cost}
The configuration of the embedding is evaluated in terms of cost, which  is the quantity minimized by $t$-SNE:

$$\sum_{i \neq j} p_{i, j} \log \frac{p_{i, j}}{q_{i, j}}$$
where $P$ and $Q$ correspond to the \textit{intra}-probability matrices without considering the interaction with the previous embedding.

\subsubsection{Distortion}
For an experiment where the items in $X^{(0)} \approx X^{(1)}$, a way to measure how well $it$-SNE places the point is to measure the distance between the initial and new position $Y^{(0)}$ and $Y^{(1)}$ and new position $Y^{(1)}$.

$$Err(Y^{(0)}, Y^{(1)}) = \sum_i |\mathbf{y}_i^{(0)} - \mathbf{y}_i^{(1)}|$$

\section{Experimental Results}
\subsection{Evolution of Gaussians}

\paragraph{Shrinking Gaussians}

This experiment reproduces the protocol proposed in \cite{10.5555/3058878.3058894},
were points are progressively attracted toward their Gaussian center of reference. The initial embedding $Y^{(0)}$ corresponding to $X^{(0)}$ is obtained with $t$-SNE, while all following embedding $Y^{(t)}$ for $t \geq 1$ are obtained with $it$-SNE using as a support the pair $(X^{(t-1)}, Y^{(t-1)})$.

\begin{figure*}[!t]
  \centering
  \includegraphics[width=1.0\textwidth]{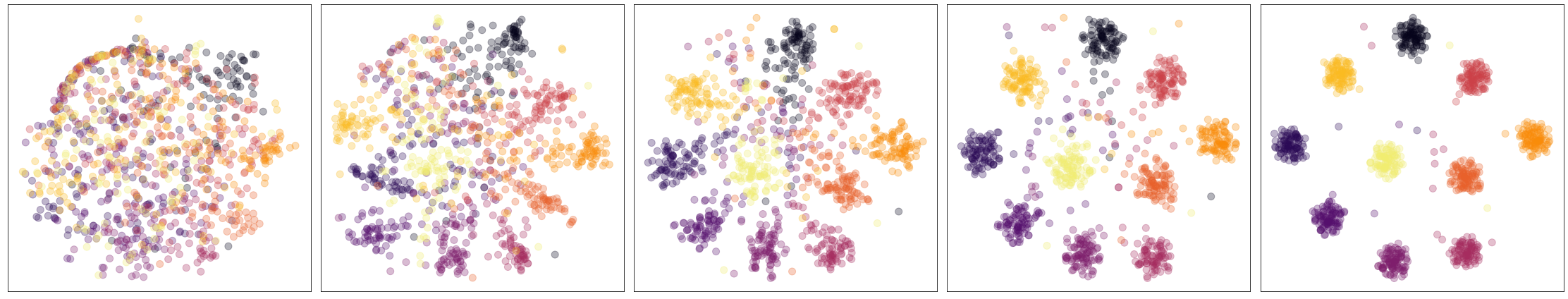}
  \caption{Shrinking Gaussians. From left to right, steps 1, 3, 5, 7 and 9 are represented. Points are colored according to their Gaussian center of reference. $\epsilon = 1.0$}
  \label{fig:gauss_shrinking}
\end{figure*}

Fig. \ref{fig:gauss_shrinking} shows the result with $it$-SNE, which transforms undistinguishable groups into well-defined groups.
Our approach works as well as $dt$-SNE, but while $dt$-SNE uses the $Y^{(t-1)}$ position to initialize $Y^{(t)}$, $it$-SNE restarts from random vectors.
This difference frees our model from restrictions on the  size and content of the dataset.
This experiment was performed on freshly generated points sampled at each time step, leading to the same results as those presented in Fig.  \ref{fig:gauss_shrinking}.

\paragraph{Growing Gaussians}

By reversing time direction, we get another set of experiments. The dataset starts with $10$ Gaussians with $\sigma(0)=0.4$, progressively increased to $\sigma(9) \approx 1.0$

\begin{figure*}[!t]
  \centering
  \includegraphics[width=1.0\textwidth]{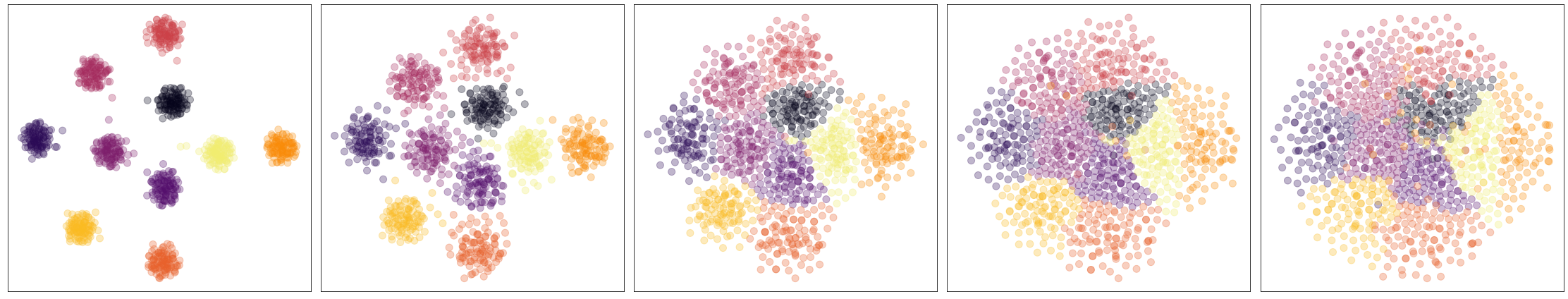}
  \caption{Growing Gaussians.
  From left to right, steps 1, 3, 5, 7 and 9 are represented.
  Points are colored according to their Gaussian center of reference.
  The penalty factor is set to $\epsilon = 1.0$}
  \label{fig:gauss_growing}
\end{figure*}

The results are represented on Fig. \ref{fig:gauss_growing}. This task is easier than the previous, as the first embedding starts with well-separated clusters.
The next embedding support is of higher quality than in the previous experiment where the variance was larger. Even after moving to a noisier dataset (the last plot of \textit{growing Gaussians} has almost the same variance as the first plot of \textit{shrinking Gaussians}), the separation between items of different clusters is preserved even if clusters are not spaced from each other.
A support embedding of good quality helps to guide items belonging to a noisy dataset, building a better embedding.

\paragraph{Change in Density}

The last visual result to present with Gaussians focuses on density changes with a fixed $\sigma$.
The number of samples per Gaussian changes at each step. For Gaussian $g$ at step  $t$, the number of items generated around $\pmb{\mu}_g$ is calculated as $n_g(t) = \lfloor n_0 r_g^t \rceil$.

\begin{figure*}[!t]
  \centering
  \includegraphics[width=1.0\textwidth]{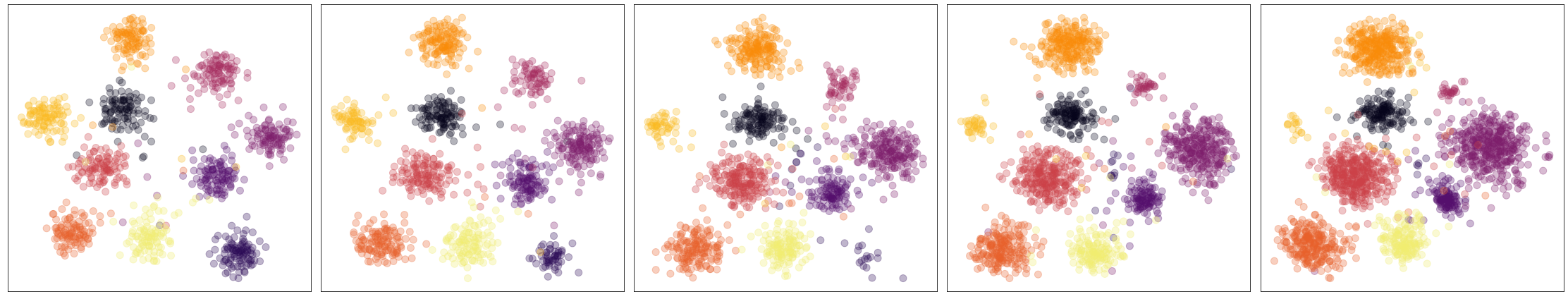}
  \caption{Evolving Gaussians.
  $10$ Gaussians of various size with $\sigma=0.4$.
  From left to right, time steps 0 to 4 are displayed.
  The penalty parameter is set to $\epsilon = 1.0$.
  Points are colored according to their Gaussian center of reference
  }
  \label{fig:gauss_density_change}
\end{figure*}

Fig. \ref{fig:gauss_density_change} illustrates the result of this process. At the start, all clusters have the same size and density and are spaced equally from each other.
As time passes, some of the Gaussians grow in size, while others shrink.
The area used by shrinking Gaussians decreases while growing Gaussians expand over the space available. The cluster positions are preserved despite the change of density unless too few items are present to allow the cluster aggregation.

\subsection{Influence of $\epsilon$}

In this subsection, we discuss the impact of $\epsilon$ on the applied forces.
For simplicity of the notation, the elements $p_{i, j}$ and $q_{i, j}$ correspond to $p^{(0, 1)}_{i, j}$ and $q^{(0, 1)}_{i, j}$,
with $i$ an element of the support dataset $(0)$ while $j$ an element of the dataset to embed $(1)$. The same simplification is applied to $y_i^{(0)}$ and $y_j^{(1)}$.

\subsubsection{Forces}
The factor $\epsilon$ has an impact on $Q$ and the gradient. The strength of the gradient is reduced, as $\epsilon$ plays in $\left(1 + \|\mathbf{y}_i - \mathbf{y}_j\| + \epsilon^2 \right)^{-1}$.
As $\epsilon$ grows, the forces coming from the support embedding vanish.

The  $\epsilon$ factor has an impact on the influence of items by changing the numerator and denominator of $Q$.
When $\epsilon$ increases, the numerator $\left(1 + \|\mathbf{y}_i - \mathbf{y}_j\| + \epsilon^2\right)^{-1}$ decreases for all item pairs. This value decays faster for items $i$ and $j$ that are close to each other than those which are not.
The denominator of $Q$ in \eqref{eq:inter_output_proba},  $V' = \sum_{i, j} \left(1 + \|\mathbf{y}_i - \mathbf{y}_j\| + \epsilon^2\right)^{-1}$ decreases when $\epsilon$ increases. As both numerator and denominator of $Q$ decrease, the effective variation depends on the item proximity. $Q$ increases for a pair of distant items and decreases for items that are close. The growth of $\epsilon$ reduces the variance, converging to  $\lim_{\epsilon \rightarrow \infty} \mathbb{V}(Q) = 0$,
and homogenizes the value of $q_{i, j}$ to $\lim_{\epsilon \rightarrow \infty} q_{i, j} = \frac{1}{n_0 n_1}$.

\begin{figure}[ht]
  \centering
  \includegraphics[width=3.5in]{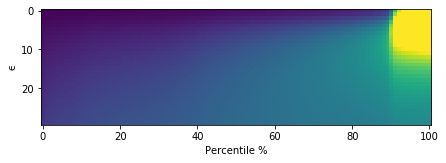}
  \caption{Percentile values of the $q_{i, j}$ for different values of $\epsilon$.
  Colors are linear with the value of $Q$:
    dark colors correspond to value close to $0$ while bright color to high value.
  The color saturates in yellow at $3 n^{-2}$}
\label{fig:gradient}
\end{figure}

Fig. \ref{fig:gradient} illustrates the evolution of $Q$ values with $\epsilon$, using the dataset with 10 Gaussians.
For the low value of $\epsilon$, only items belonging to the same Gaussian interact together (i.e., 10\% of the points). As $\epsilon$ grows, the weights of the nearest neighbors decrease to the profit of more distant neighbors.

Because of the kernel asymmetry between $P$ and $Q$, the reduction of variance and the convergence to the mean of $Q$ leads to two different behaviors, depending on the closeness of two items.
The items are divided into two classes based on the value of $p_{j , i}$. The closer neighbors with $p_{j , i} > \frac{1}{n_0 n_1}$ are considered as the nearest neighbors while the other distant neighbors.
For nearest neighbors, the artificial distancing created by $\epsilon$ lowers the output probability $q > q^{\epsilon}$, for $q^{\epsilon}=q(\epsilon > 0)$ and $q=q(\epsilon=0)$. $P$ and $Q$'s difference is $p - q^{\epsilon} > p - q$ is then larger, which leads to larger attractive forces from the close neighborhood.
The opposite effect happens for distant neighbors where $q < q^{\epsilon}$, which leads to $p - q > p - q^{\epsilon}$, generating repulsive forces.

To summarize, on the one hand, the forces are globally lowered as $\epsilon$ impacts the gradient,
and on the other hand, the discrimination between nearest and distant neighbors grows as $\epsilon$ amplifies the asymmetry.

\subsubsection{Displacement from Origin}

A way to study the two contributions can be done by looking at the ability of $it$-SNE to recover the exact embedding positions. For a dataset $X^{(0)}$, first is computed $Y^{(0)}$ with:

$$Y^{(0)}\leftarrow t\text{-SNE}(X^{(0)})$$

followed by:

$$Y^{(1)}\leftarrow it\text{-SNE}(X^{(0)}, X^{(0)}, Y^{(0)})$$

\begin{figure}[ht]
  \centering
  \includegraphics[width=0.8\columnwidth]{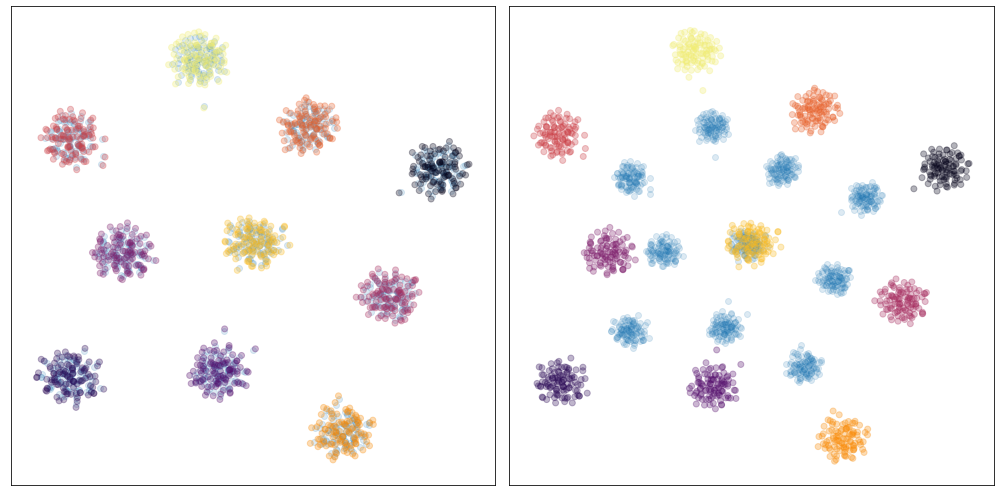}
  \caption{Mapping of Gaussians with standard deviation $\sigma = 0.4$.
  The embedding $Y^{(0)}$ is colored in light blue on each plot,
  while items of $Y^{(1)}$ are colored according to their cluster of reference.
  Left: $\epsilon=1$, right: $\epsilon=25$
  }
\label{fig:approximate_position}
\end{figure}

Fig. \ref{fig:approximate_position} shows the cluster conformation for two different $\epsilon$.
The clusters are correctly matched, both in classes and in positions for $\epsilon = 1$, as $Y^{(1)}$ masks $Y^{(0)}$.
However, for $\epsilon=25$, while the cluster classes are correctly matched, they are distant from the original position.

To study the transition between the two conformations, we look at the distortion $Err(Y^{(0)}, Y^{(1)})$, which measures the distance between the initial and final position.

\begin{figure}[ht]
  \centering
  \includegraphics[width=0.5\textwidth]{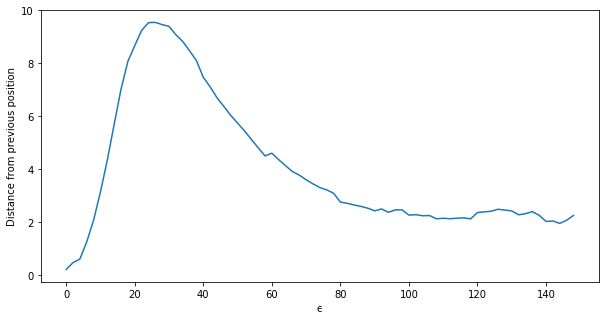}
  \caption{Evolution of the embedding error $Err(Y^{(0)}, Y^{(1)})$ with $\epsilon$ using 10 Gaussians with standard deviation $\sigma=0.4$
  }
\label{fig:exact_distance}
\end{figure}

Fig. \ref{fig:exact_distance} shows the impact of an increase of $\epsilon$ over the item locations.
For low $\epsilon$, the error is almost $0$. The average distance error for $\epsilon = 0 $ is $0.204$, while the average distance to the first nearest neighbor is $0.218$, for an embedding of diameter $32.1$.
With increasing values of $\epsilon$, the distances between initial and final positions grow. The right part of Fig. \ref{fig:approximate_position} illustrates the situation. The repulsive forces between distant neighbors are stronger than the attraction of the nearest neighbors. This pushes the clusters even further away from each other, increasing the distance from the initial position.
The maximal distortion is reached around $\epsilon = 25$, a value of the order of the embedding diameter.

After this maximum, the distortion error decreases to stabilize at a value of $2.6$.
Compared to the diameter of one cluster presented in \ref{fig:approximate_position}, the value is relatively similar, around $2.88$.
The clusters overlap, but the forces are not strong enough to accurately bring them to their exact location.

\subsubsection{Cost}
$it$-SNE tries to minimize two costs at the same time. They might not be compatible, with opposite gradient directions. The intra cost allows assessing the  embedding quality to see if the embedding could reach a correct minimum, or if the inter forces constrained the embedding to a non-optimal state.

\begin{figure}[ht]
  \centering
  \includegraphics[width=0.5\textwidth]{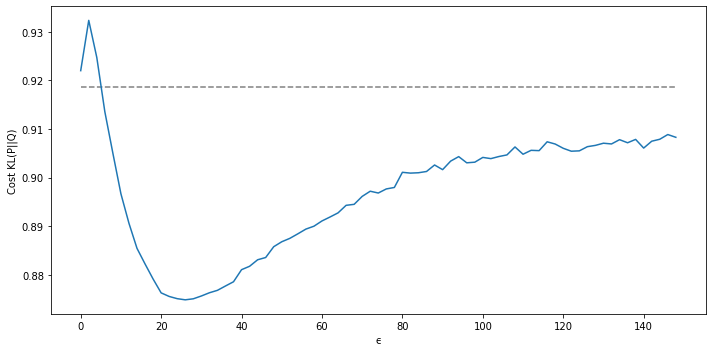}
  \caption{Evolution of Kullback-Leibler cost with increasing $\epsilon$.
  Gaussians with standard deviation $\sigma = 0.4$.
  The dashed line corresponds to the cost of embedding $Y^{(0)}$ obtained with the regular $t$-SNE}
\label{fig:cost_change}
\end{figure}

Fig. \ref{fig:cost_change} presents the results for the same conformations as in Fig. \ref{fig:exact_distance}.
The cost is slightly higher to start with, but it decreases as $\epsilon$ increases to arrive at a local cost minimum. This local minimum corresponds to the distortion maxima of Fig. \ref{fig:exact_distance}. The conformation $\epsilon=25$ is a more stable configuration than the initial one. For larger values, the forces' strength  decrease, but stay below the baseline cost, corresponding to the support configuration.
It is to note that the cost difference between the lowest and highest cost value in Fig. \ref{fig:cost_change} is relatively small.
All conformations are relatively good, some a little bit more than others.

\subsubsection{Convergence Speed}

In our protocol, the number of training step was fixed to control the computational time.
For a support embedding of $n_0$ items and a new dataset to embed of $n_1$ items, the $t$-SNE cost is proportional to $n_1^2$, while the cost for $it$-SNE is proportional to $n_1^2 + n_0 n_1$. In our experiments $n_0 = n_1 = 1000$ which means that the number of operations performed by $it$-SNE is twice the number of $t$-SNE.

\begin{figure}[ht]
  \centering
  \includegraphics[width=0.7\textwidth]{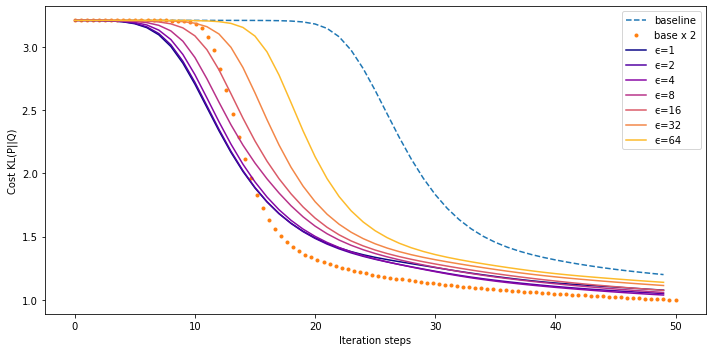}
  \caption{Evolution of the Kullback-Leibler cost over training time,
  with 10 Gaussians of standard deviation $\sigma = 0.4$.
  The yellow to dark lines correspond to $it$-SNE for several value of $\epsilon$
  The blue dashed line corresponds to the cost evolution for $t$-SNE.
  The dotted line corresponds to baseline with $t$-SNE, speeded by a factor $2$
  }
\label{fig:cost_time}
\end{figure}

Fig. \ref{fig:cost_time} shows the cost evolution for several values of $\epsilon$.
All curves start with a plateau, which corresponds to when the learning rate is not boosted enough to lead to significant changes per steps. The lower $\epsilon$ is, the shorter the time spent on the plateau is. A decay part follows the plateau, which smoothly slowed down until convergence.

The dotted orange line allows comparing $t$-SNE with $it$-SNE on the number of computational operations.
$t$-SNE is faster than $it$-SNE, but the difference between the two is not very large.

\subsection{Citation Graph Embedding}

Gaussian clusters are easy to study as the parameters are fully controlled.
Citation graphs are selected to illustrate a real-world example of $it$-SNE capabilities.

Because a scientific article refers to relevant papers in its field, this type of dataset presents local community structures.
The evolution of the number of researchers is leading to an expansion of knowledge in many fields, and the creation of new ones. Other fields tend to disappear, because of a lack of support from the scientific community or an absence of new discoveries.
The study of these phenomena allows to retrace history and reconstruct the science phylogeny \cite{science_phylogeny}.

\begin{figure}[ht]
  \centering
  \includegraphics[width=0.8\textwidth]{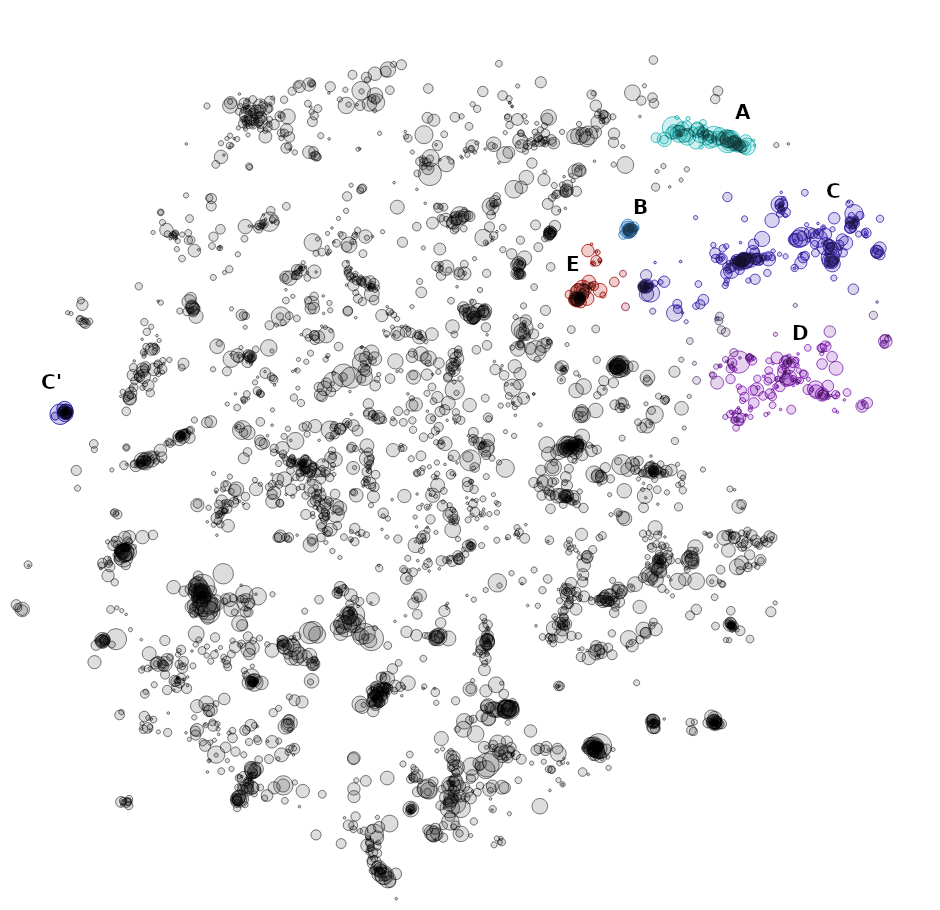}

  \caption{Citation graph of cryptographic papers in 2010, with some highlighted clusters.
    $A$: Hashing, $B$: Network Code, $C$ and $C'$: Biometry, $D$: Watermarking/Data Hiding, $E$: Passwords.
    The item size is proportional to the number of citations
    }
  \label{fig:cit_graph_global}
\end{figure}

Fig. \ref{fig:cit_graph_global} represents the papers published in the cryptography/security field in 2010.
Papers group together to form connected clusters with various shapes, sizes and densities. To the default of a clustering algorithm, some groups have been highlighted and labelled by hand with the help of documents' title and keywords.

Cluster $A$, with \textit{Hashing}'s general topic, is compact with items well connected to each others.
Cluster $D$ about \textit{Watermark} is more diffuse than $A$, but items are still grouped together. The \textit{Biometry} cluster $C$ is composed of several sub-units of various density. Cluster $B$ and $E$ about \textit{Network Code} and \textit{Password} respectively are much more compact than the other presented.

The papers in each cluster are mostly in phase with the general cluster topic, showing local unity.
Nonetheless, the reverse is not valid: a cluster about a particular topic does not enclose all documents related to it. A keyword may correspond to two different ideas, or two distinct communities may work on different aspects of the problem.
For instance, this is the \textit{Biometry} field case, which occurs twice in various embedding locations. There is one large cluster on the right and a smaller one on the left (denoted $C$ and $C'$ respectively). While the large cluster $C$ is about general biometric recognition methods, the $C'$ is focused on authentication scheme,  mixing \textit{Biometry}, \textit{Password} and \textit{Authentication} topics together.

Scientific communities are dynamic and adapt to new trends. Fields emerge, grow, interact, split, and disappear. $it$-SNE allows reusing these initial cluster positions for the next subsequent embedding, which allows tracking such dynamics.

\begin{figure}[ht]
  \centering
  \includegraphics[width=1.0\textwidth]{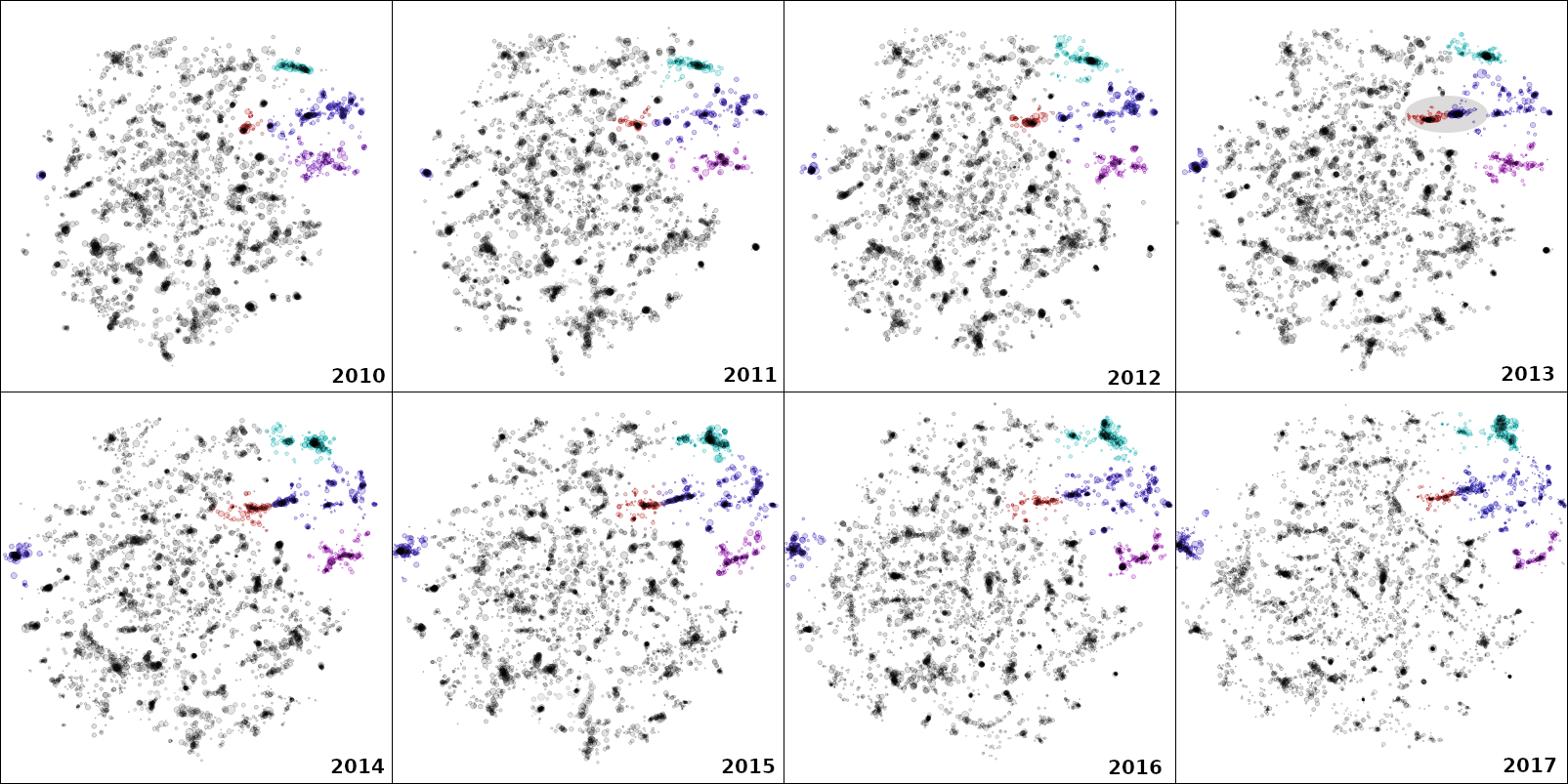}

  \caption{Citation graph of cryptographic papers, published between 2010 to 2017.
  Clusters are highlighted according to the previous figure coloring scheme.
  Size of points is proportional to the logarithmic number of citations.
  Dark areas correspond to highly grouped papers.
  The fusion between Biometry and Password clusters in 2013 is shaded in grey
  }
  \label{fig:cit_graph_evolution}
\end{figure}

Fig. \ref{fig:cit_graph_evolution} presents the evolution of the embedding \ref{fig:cit_graph_global} from 2010 to 2017, with the same color highlighting.
The general shape of the embedding is stable over time, with clusters' positions preserved.
A small drift of the clusters occurs, which is noticeable after a few embedding steps. The growth and shrink of some clusters is visible, such as for the second biometry cluster on the left which expands over the years.
Merges are also visible, such as for the \textit{Biometry} and \textit{Password} clusters, which first merge in 2013.

\begin{figure}[ht]
  \centering
  \includegraphics[width=1.0\textwidth]{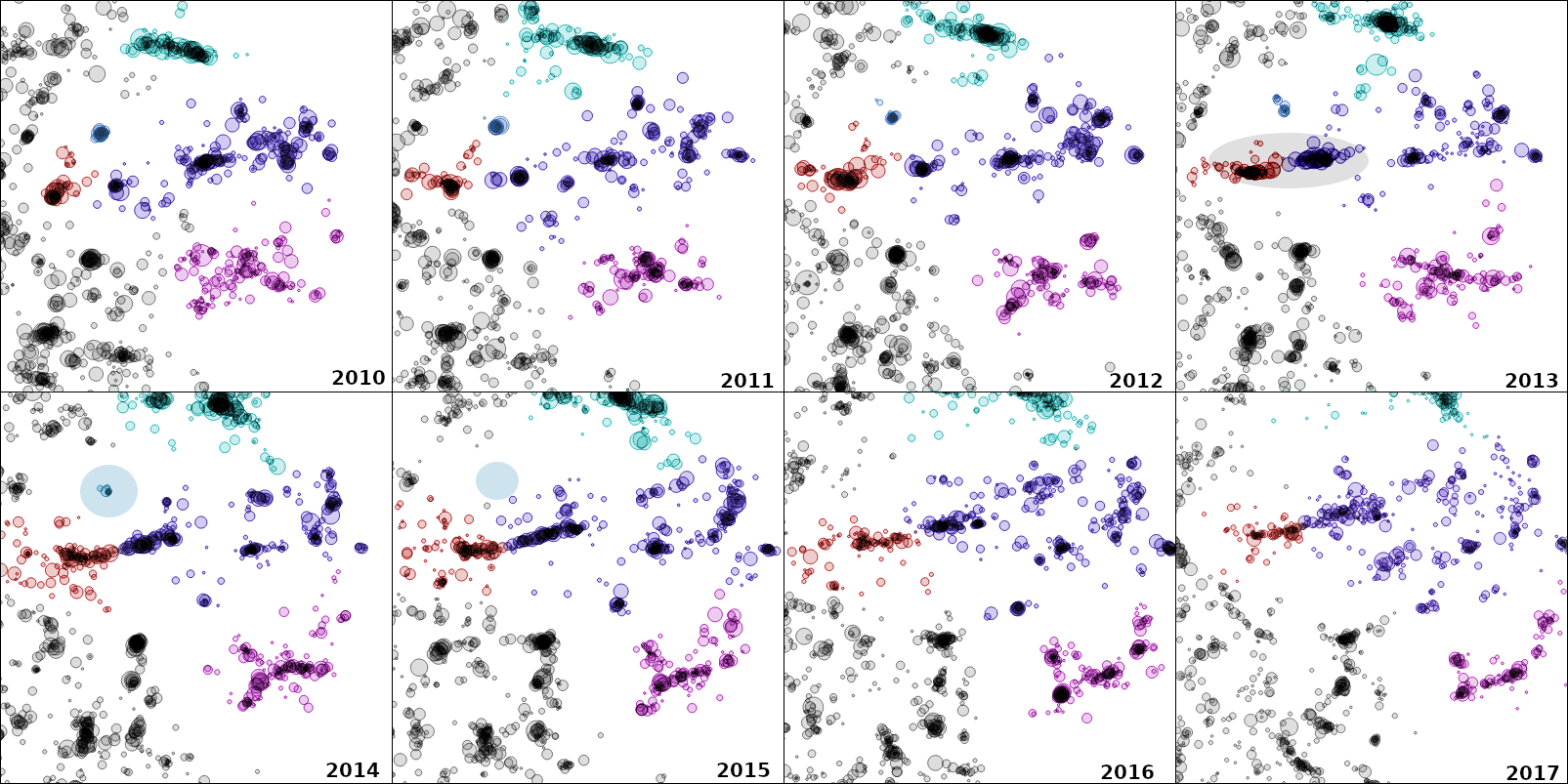}

  \caption{Citation graph of cryptographic papers, published between 2010 to 2017.
  Clusters are highlighted according to the previous figure coloring scheme.
  The size of points is proportional to the logarithmic number of citations.
  Dark areas correspond to highly grouped papers.
  The fusion between Biometry and Password clusters in 2013 is shaded in grey
 }
  \label{fig:cit_graph_evolution_zoom}
\end{figure}

To have a better view of the evolution, Fig. \ref{fig:cit_graph_evolution_zoom} shows an enlarged view of the top right area of the embedding, where \textit{Biometry} and \textit{Password} field of study clusters are.
In this area, different types of phenomena occur.
There are stable clusters present, such as the \textit{Hashing},  \textit{Biometry}, and \textit{Watermark}, with constant size and density.
\textit{Password} cluster growths in size, while \textit{Network Coding} disappeared in 2015. The \textit{Biometry} and \textit{Password} groups have been interacting with each other and began merging in 2017.

As the clusters have been human extracted, no metric measure has been tested.
Tables listing the most cited paper for each year for the different clusters
 are presented in the appendix (Tables \ref{tab:cluster_hashing}, \ref{tab:cluster_network_coding}, \ref{tab:cluster_biometry},
\ref{tab:cluster_biometry_bis}, \ref{tab:cluster_watermark}, \ref{tab:cluster_password}).
The articles presented in these tables are very consistent with the theme of the cluster from which they originate.

\section{Discussion}

\subsection{Complexity}
The normal complexity of $t$-SNE is in $\mathcal{O}(n^2)$, where $n$ is the number of items.
While some optimization exists when the dataset is tabular, like the Barnes-Hut optimization \cite{journals/corr/abs-1301-3342} which reduces the cost to $\mathcal{O}(n \log n)$, for data like graphs from which a distance matrix can be obtained, this quadratic cost is prohibitive.

If the dataset is cut into $k$ equivalent pieces of $m = \frac{n}{k}$ pieces, the algorithm would run in $\mathcal{O}(2km^2) =\mathcal{O}(\frac{2}{k}n^2)$, where the $2 \times$ stands for the two gradient parts. This reduces by a factor $\frac{k}{2}$ the complexity.
The complexity here measures the average number of operations for one run of an iteration step. However, as the algorithm uses gradient descent, a convergence criterion governs the total number of steps. Intuitively, the number of steps required to converge for a large dataset seems larger than for a smaller dataset.
The decomposition of the dataset into pieces would reduce the effective computational time.

Concerning the memory requirements, the normal $t$-SNE requires a storage space of $2n^2$, necessary for the matrix $P$ and $Q$. Using $it$-SNE with a dataset split into $k$ pieces, $4$ matrix of size $m^2$,  $(P^{(t)}, Q^{(t)}, P^{(t-1,t)}, Q^{(t-1,t)})$ are needed to compute the embedding.
The total amount of memory required is then $4m^2 = 4 \left(\frac{n}{k}\right)^2$, which is more interesting, as $\frac{k^2}{2}$ reduces its cost. The computational time can be extended on a machine, but not its memory. Our proposed method can be helpful as way to map large datasets by cutting them into smaller pieces.

\subsection{Selecting $\epsilon$}

\subsubsection{Speedup}
The parameter $\epsilon$ controls the applied forces and the gradient strength.
A low value of $\epsilon$ creates strong forces which leads to fast convergence.
Nonetheless, forces prevent items to move to other locations. An increase of $\epsilon$ would relax the system and help an item to arrive on a low energy state.
As for the \textit{early exaggeration} trick, it would be benefical to start with a small value of $\epsilon$ and then finish with a larger value.

\subsubsection{$\epsilon$ and Perplexity}
The perplexity governs the number of neighbors taken into account. The numerator of $P$ in equations \eqref{eq:input_proba} and  \eqref{eq:inter_input_proba} grows with a perplexity increase for all items.
The denominator grows too, and leads to a decrease of $P$ for the nearest neighbors.
It affects on $Q$ which needs to decrease. The distances between $Y$'s increase with larger perplexity.

If the support embedding has a different perplexity than the target perplexity, a large $\epsilon$ may help to adapt to the new perplexity, by relaxing forces strength.
The clusters would be attracted to nearby position, and the intra forces would arrange the local shape.

\subsection{Adaptation to Large Changes of Density}

Our experiments have been done with temporal datasets with constant or slowly evolving size.
The use of a support dataset of highly different size may constrain the system optimization. For a fixed perplexity, the diameter of an embedding grows with the size of the dataset  in $(n)^{\frac{1}{d_e}}$. Two datasets of different size would have a radius of $r^{(0)}$ and $r^{(1)}$. The items located in the middle of the embedding can be correctly matched with the support items. However, for peripheral items, there would be a gap of $|r_0 - r_1|$.
$it$-SNE would create a distortion, constraining items of $(1)$ to expand if $r_0 > r_1$, which would increase the interdistances between items in $Y^{(1)}$. For $r_0 < r_1$, a shrinking would occur leading to the same distortion. As intra forces of $(0)$ do not play any role, a trick to free from this constraint would be to scale $Y^{(0)}$ into $Y^{(0*)}$, using the scaling ratio $s = \left(\frac{n_1}{n_0}\right)^{\frac{1}{d_e}}$, such as $Y^{(0*)} = sY^{(0)}$.
This would help for datasets of similar densities. For different densities, local distortions would still occur.

\subsection{Using More than One Support Embedding}

The method proposed to use one embedding to support the generation of a new one.
A natural question arises about the possibility of using the support of two or more embeddings. This case may happen if two embeddings for $t_0 < t_2$ have been obtained but not for $t_1$ yet, with $t_0 < t_1 < t_2$. Intuitively, the constrains engendered by the two embeddings might be equivalent to a single one.
If more embeddings were to be taken into account, all grouped embedding cost must not prevent the intra forces from playing their role.

For a dataset $X^{(k)}$ taking support of datasets $\mathcal{X} = \{X^{(i)}\}_{i=1:k-1}$ with respective embeddings $\mathcal{Y} = \{Y^{(i)}\}_{i=1:k-1}$, the cost could be rewritten as:

$$C^{(k)}_{tot} = C^{(k)} + \frac{1}{k-1}\sum_{i=1}^{k-1} C^{(i, k)}$$

This adaptation allows generating embedding in between two existing embeddings.
Another use of this adaptation is the multivariate case, like for geolocation coordinates,
where the new embedding might take the support using multiple non-equivalent datasets. For  $\mathcal{X} = \{X^{(i)}\}_{i=1:k-1}$ with relative importance $W = \{w_i | w_i > 0\}_{i=1:k-1}$, the weighted cost would have the form:

$$C^{(k)}_{tot} = C^{(k)} + \frac{1}{\sum_{i=1}^{k-1} w_i}\sum_{i=1}^{k-1} w_i C^{(i, k)}$$

Note that the use of multiple support embeddings increases the cost linearly with the total number of items.
Nonetheless, if the nearest datasets are too small to serve as a support, the use of more than one embedding may helps to preserve embedding knowledge and enhance long term coherency.

\subsection{Binary Computation}
To create several coherent embeddings for a succession of datasets, the process can be speeded-up by distributing the embedding tasks.
If there are $k$ datasets to embed, instead of computing the embeddings in a sequential way, using the support of $t-1$ to compute $t$, the use of another more distant support would help. The closer the support, the better it would be, as the distribution difference between two neighbor datasets is expected to be lower than for distant datasets.

The computation starts with an initial embedding for $\lfloor \frac{k}{2} \rceil$.
Then, the left and right intervals are divided in their middle. An embedding is issued for $\lfloor \frac{k}{4} \rceil$ and $\lfloor \frac{3k}{4} \rceil$.
Then, the embedding for subset $\lfloor \frac{1}{8} k \rceil$ can be computed using the support of $\lfloor \frac{k}{4} \rceil$,
and $\lfloor \frac{7k}{8} \rceil$ using the support of $\lfloor \frac{3k}{4} \rceil$. For the middle parts $\lfloor \frac{3}{8} k \rceil$ and $\lfloor \frac{5}{8}k \rceil$, their respective embedding is computed using two support embeddings, respectively using $(\lfloor \frac{k}{4} \rceil, \lfloor \frac{k}{2} \rceil)$ and  $(\lfloor \frac{k}{2} \rceil, \lfloor \frac{3k}{4} \rceil)$.
The procedure is repeated recursively until all embeddings have been obtained. This decomposition allows to speedup the process from $\mathcal{O}(k)$ to $\mathcal{O}(\log_2(k))$.

\section{Conclusion}
This paper presents a method adapting $t$-SNE algorithm to reuse a previous embedding to generate a new one.
Compared to the base method $t$-SNE, an additional cost term is added. This cost links the new items to embed to the support embedding, creating attractive forces.
These forces enable the similar items from the support and current datasets to be located on the same embedding area. Clusters are coherent in the location from one embedding to the other, enabling the reuse of a classification algorithm on both.

$it$-SNE was tested on two datasets. The first used synthetic Gaussians forming dense clusters, evolving in density and size over time.
The second was the scientific citation graph restricted to cryptography related papers, with small, sparse communities. The algorithm was successful at preserving the cluster locations in both experiments, while preserving $t$-SNE embedding aspect.

Compared to $t$-SNE, the computational complexity and the memory requirement of $it$-SNE are doubled.
Nonetheless, the use of a support embedding speedups the convergence process of $it$-SNE. The total number of operations of $it$-SNE is, in practice, equivalent to $t$-SNE.

We proposed two extensions: the first to the multivariate case and the second to distribute the embedding computation.
One unsolved problem yet is the adaptation to highly different densities, as $t$-SNE mechanism tries to keep average distance between neighbors constant,
which leads to an expansion of the embedding with increasing dataset size.

$it$-SNE can be used for many purposes, such as monitoring, anomaly detection, network analysis, allowing to track the evolution of clusters in a low dimensional space.
The method is not restricted to temporal datasets and could be used to study the impact one variable's impact on the dataset distribution.

\bibliographystyle{unsrtnat}
\bibliography{references}

\section*{Appendix: Most Cited Papers}

Tables \ref{tab:cluster_hashing} to \ref{tab:cluster_password} list the most cited document per year per topic.

\begin{table}[ht]
  \caption{Hashing}
  \begin{tabular}[t]{llc}
    \hline
     Year & Title  \\
    \hline
    2010 & Semi-supervised Hashing for Scalable Image Retrieval \\
    2011 & Minimal Loss Hashing for Compact Binary Codes \\
    2012 & Image Signature: Highlighting Sparse Salient Regions  \\
    2013 & Inter-media Hashing for Large Scale Retrieval from Heterogeneous data sources\\
    2014 & Supervised Hashing for Image Retrieval via Representation Learning \\
    2015 & Supervised Discrete Hashing \\
    2016 & Deep Supervised Hashing for Fast Image Retrieval  \\
    2017 & Learning Discriminative Binary Codes for Large-scale Cross-modal Retrieval\\
    \hline
  \end{tabular}
  \label{tab:cluster_hashing}
\end{table}

\begin{table}[ht]
  \caption{Network Coding}
  \begin{tabular}[t]{llc}
    \hline
     Year & Title \\
    \hline
    2010 & Secure network coding over the integers \\
    2011 &  Secure Network Coding on a Wiretap Network \\
    2012 &  Cooperative Defence Against Pollution Attacks in Network Coring Using SpaceMac \\
    2013 &  An Efficient Homomorphic MAC with Small key Size for Authentication in Network Coding \\
    2014 &  A Lightweight Encryption Scheme for Network-Coded Mobile Ad Hoc Networks\\
    \hline
  \end{tabular}
  \label{tab:cluster_network_coding}
\end{table}

\begin{table}[ht]
  \caption{Biometry}
  \begin{tabular}[t]{llc}
    \hline
     Year & Title \\
    \hline
    2010 & Unobtrusive User-Auth on Mobile Phone Using Biometric Gait Recognition\\
    2011 &  A survey on Biometric Cryptosystems and cancelable biometrics \\
    2012 &  Touch me once and I Know it's you \\
    2013 &  Touchalytics: On the Applicability of Touchscreen Input as a Behavioral Biometric for Continuous Authentication \\
    2014 &  Image quality Assessment for Fake Biometric Detection  \\
    2015 &  Deep Representation for Iris, Face and Fingerprint Spoofing Detection  \\
    2016 &  Continuous User Authentication on Mobile Devices  \\
    2017 &  MagNet: A Two-Pronged Defense against Adversarial Examples \\
    \hline
  \end{tabular}
  \label{tab:cluster_biometry}
\end{table}

\begin{table}[ht]
  \caption{Biometry and Authentication Schemes}
  \begin{tabular}[t]{llc}
    \hline
     Year & Title \\
    \hline
    2010 & An Efficient Biometrics-based Remote User Authentication Scheme using Smart Cards \\
    2011 & Cryptanalysis and Improvement of a Biometrics-based Remote  user Authentication Scheme using Smart Cards\\
    2012 & A secure Authentication Scheme for Telecare  Medicine Information Systems \\
    2013 & A Temporal-Credential-Based Mutual Authentication and key agreement scheme for wireless sensor networks\\
    2014 & A Novel User Authentication and Key Agreement Scheme for heterogeneous ad hoc wireless sensor networks, \\
		& based on the Internet of Things notion\\
    2015 & Robust Biometrics-Based Authentication Scheme for Multiserver Environment \\
    2016 & An efficient User Authentication and Key Agreement Scheme for heterogeneous wireless sensor network \\
		& tailored for the Internet of Things environment\\
    2017 & Anonymous Authentication for Wireless Body Area Networks With Provable Security \\
    \hline
  \end{tabular}
  \label{tab:cluster_biometry_bis}
\end{table}

\begin{table}[ht]
  \caption{Watermark}
  \begin{tabular}[t]{llc}
    \hline
     Year & Title  \\
    \hline
    2010 & Review: Digital Image Steganography:  Survey and analysis of current methods \\
    2011 & Reversible Data Hiding in Encrypted Image \\
    2012 & Separable Reversible Data Hiding in Encrypted Image \\
    2013 & Digital Image Forgery Detection using Passive Techniques  \\
    2014 & Reversibility improved data Hiding in Encrypted images  \\
    2015 & RAISE: a Raw Images Dataset for Digital Image Forensics  \\
    2016 & Reversible Data Hiding: Advances in the Past Two Decades \\
    2017 & Fragile Image Watermarking with Pixel-wise Recovery based on Overlapping Embedding Strategy \\
    \hline
  \end{tabular}
  \label{tab:cluster_watermark}
\end{table}

\begin{table}[ht]
  \caption{Password}
  \begin{tabular}[t]{llc}
    \hline
     Year & Title  \\
    \hline
    2010 & Encountering Stronger Password Requirements: user attitudes and behaviors  \\
    2011 &  Of Passwords and People: Measuring the Effect of Password-composition policies  \\
    2012 &  The Quest to Replace Passwords: a Framework for Comparative Evaluation of Web Authentication Schemes \\
    2013 &  Patterns in the Wild: a Field Study of the Usability of Pattern and pin-based authentication on mobile devices\\
    2014 &  It's a Hard Lock Life: A Field Study of Smartphone (Un)Locking Behavior and Risk Perception\\
    2015 &  "... No one Can Hack My Mind": Comparing Expert and Non-Expert Security Practices\\
    2016 &  Who are you? A Statistical Approach to Measuring User Authenticity \\
    2017 &  Zipf's Law in Passwords\\
    \hline
  \end{tabular}
  \label{tab:cluster_password}
\end{table}

\end{document}